\pdfoutput=1

\documentclass[11pt]{article}

\usepackage[]{EMNLP2022}

\usepackage{times}
\usepackage{latexsym}

\usepackage[T1]{fontenc}

\usepackage[utf8]{inputenc}

\usepackage{microtype}

\usepackage{inconsolata}

\usepackage{microtype}

\usepackage{amsmath,amsfonts,bm}

\def\eqref#1{equation~\ref{#1}}

\def\1{\bm{1}}

\DeclareMathAlphabet{\mathsfit}{\encodingdefault}{\sfdefault}{m}{sl}
\SetMathAlphabet{\mathsfit}{bold}{\encodingdefault}{\sfdefault}{bx}{n}

\usepackage{mathabx}
\usepackage{subfig}
\usepackage{float}
\usepackage{bbm}
\usepackage{cleveref}

\usepackage{times}
\usepackage{soul}
\usepackage{url}
\usepackage{hyperref}
\usepackage[utf8]{inputenc}
\usepackage{caption}
\usepackage{graphicx}
\usepackage{multirow}
\usepackage{amsmath}
\usepackage{amsthm}
\usepackage{booktabs}
\usepackage{algorithm}
\usepackage{algorithmic}
\usepackage[normalem]{ulem}
\useunder{\uline}{\ul}{}
\usepackage{tabularx}
\usepackage{listings}
\usepackage{filecontents} 
\usepackage{multicol}
\title{PCFG-based Natural Language Interface Improves Generalization for Controlled Text Generation}

\author{Jingyu Zhang \\
  Johns Hopkins University \\
  \texttt{jzhan237@jhu.edu} \\\And
  James Glass \\
  MIT \\
  \texttt{glass@mit.edu} \\\And
  Tianxing He \\
  University of Washington \\
  \texttt{goosehe@cs.washington.edu} \\}

\begin{document}
\maketitle
\begin{abstract}
Existing work on controlled text generation (CTG) assumes a control interface of categorical attributes. In this work, we propose a natural language (NL) interface, where we craft a PCFG to embed the control attributes into natural language commands, and propose variants of existing CTG models that take commands as input. 
In our experiments, we design tailored setups to test model's generalization abilities. We find our PCFG-based command generation approach is effective for handling unseen commands compared to fix-set templates; our proposed NL models can effectively generalize to unseen attributes, a new ability enabled by the NL interface, as well as unseen attribute combinations. Interestingly, we discover that the simple conditional generation approach, enhanced with our proposed NL interface, is a strong baseline in those challenging settings.

\end{abstract}

\section{Introduction}
\label{sec:intro}

With the advancement of large scale pretraining, language models (LM) are now able to generate increasingly more realistic text \citep{gpt2_radford2019language,gpt3_NEURIPS2020_1457c0d6, DBLP:journals/corr/abs-2112-11446, Hoffmann2022TrainingCL, Smith2022UsingDA, Thoppilan2022LaMDALM}. Therefore, how to control the generation of LMs have become an important research topic. 
In \textit{controlled text generation} (CTG), a series of works \citep[\textit{inter alia}]{nitish2019ctrl, Dathathri2020Plug,krause-etal-2021-gedi-generative,yang-klein-2021-fudge,liu-etal-2021-dexperts,yu-etal-2021-attribute-alignment,Li2022DiffusionLMIC} propose model frameworks to generate text conditioned on some desired (user-specified) attribute $a$. These attributes, which depend on the datasets of interest, could be topic, formality, sentiment, etc. 


\begin{figure}[t]
    \centering
    \includegraphics[width=0.85\linewidth]{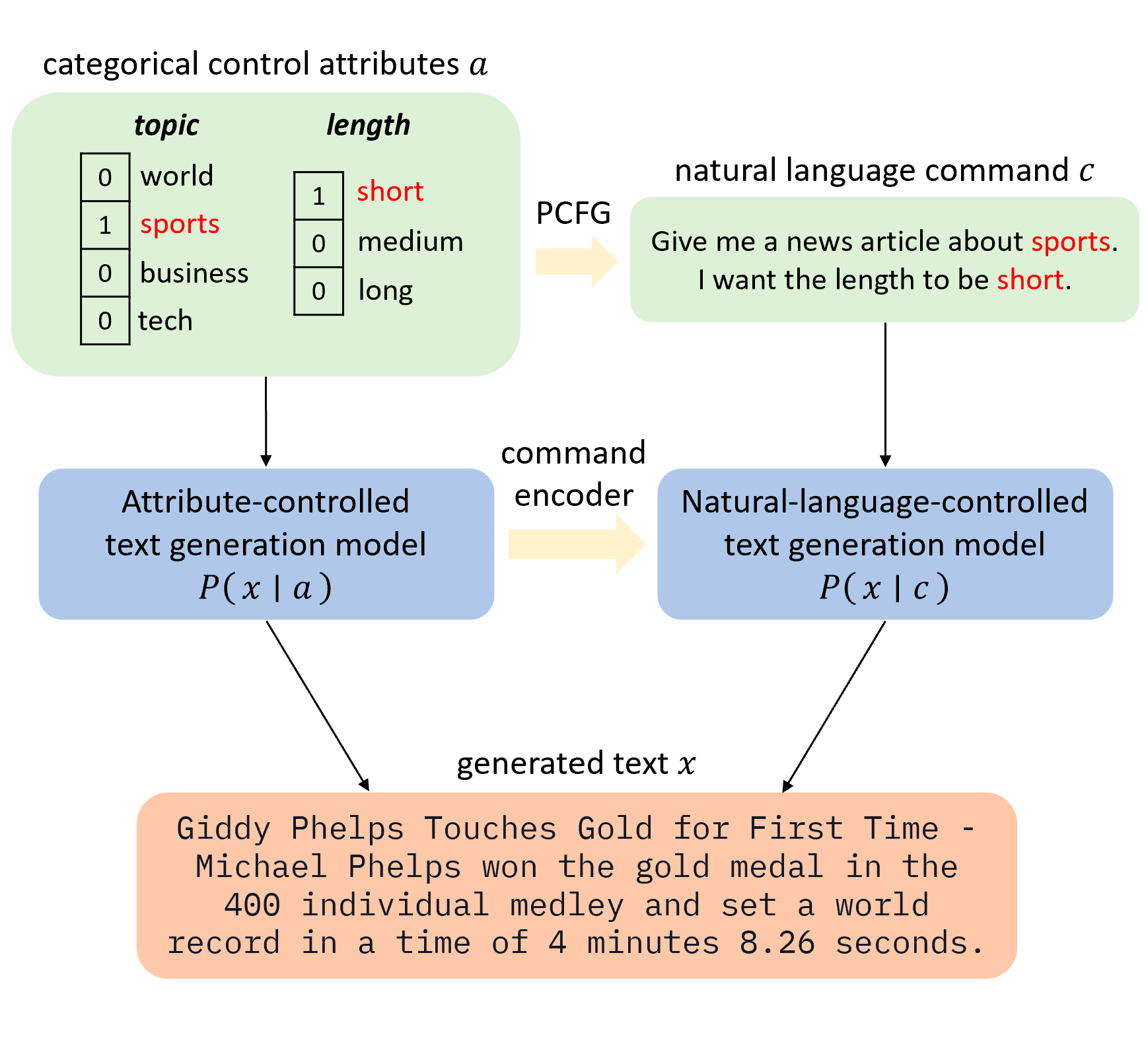}
    \caption{We use PCFG to embed categorical control attributes into natural language command. Correspondingly, we propose generation models that take command as input.}
    \label{fig:mo_interface}
\end{figure}

An important assumption behind this controlled generation setting is that the attributes are chosen from \textbf{a fixed set} (i.e., they are treated as categorical random variables). Although this setting is convenient, it seriously limits the applications of the CTG system: (1) Since the attribute set is fixed during training, it would be impossible for the model to generalize to unseen options if used as-is.
(2) This interface is not very human-friendly, because it could be difficult for users to navigate through the (possibly long) lists of options.
Motivated by these limitations, in this work we propose a \textit{natural language interface} for CTG, illustrated in \autoref{fig:mo_interface}. With this change of interface, the input to the CTG model changes from one-hot vectors to natural language commands (for short, \textit{commands})
To efficiently train this system and enable it to generalize, we design a probabilistic context-free grammar (PCFG) to embed categorical attributes into a diverse set of natural language commands. 

Using natural language instruction has been explored in recent work \citep{Sanh2021MultitaskPT, wei2022finetuned, Mishra2022CrossTaskGV, reif-etal-2022-recipe, schick-schutze-2021-shot}.
Our work differs from them in: (1) We focus on the task of CTG as opposed to the performance on cross-task generalization, and design tailored scenarios for evaluation. (2) We introduce PCFG for command generation, which has not been explored by previous work. We discuss this relationship in more details in Section \ref{sec:related}.

The change of interface brings several immediate benefits: 
(1) Natural language inputs enable the system to generalize to unseen attribute options (as long as it can be expressed in natural language). 
(2) Unlike fixed-set template sentences in previous works, the PCFG can generate diverse natural language variation during training, which we will show is crucial for generalization.
(3) The input process become more natural and interactive to a human user, and it can be linked with, for example, a speech recognition module. 

With this new interface we propose variants of several existing CTG systems that take commands as input, and design experiments to compare different CTG models under tailored scenarios. We briefly summarize our main contributions below:
\begin{itemize}
    \item We propose a PCFG-based natural language interface for controlled text generation. The natural language interface enables zero-shot generalization on control attributes unseen during training, a capability previously impossible due to the fixed-set assumption.
    \item We show that training with commands generated by a PCFG is an effective method for increasing natural language variation over using fixed-set templates, allowing natural language CTG models to better generalize to commands unseen during training.
    \item We test the proposed natural language CTG models on settings where the models need to generalize to unseen attributes and attribute combinations. Surprisingly, the simple conditional generation approach is shown to be a strong baseline in these challenging setups.

\end{itemize}

\vspace{0.5mm}
\section{Related Work}
\label{sec:related}

\paragraph{Instruction Following} A recent series of work proposes to describe NLP tasks in natural language, and use the task description as an instruction to promote zero-shot generalization for LMs \citep[\textit{inter alia}]{Sanh2021MultitaskPT, wei2022finetuned}. Such task descriptions are manually created, detailed definitions of NLP tasks, which contain explanations about input, output, emphasis, and possibly a small number of demonstrative examples. InstructGPT \citep{Ouyang2022TrainingLM} uses an RL policy to improve LM's capability to follow user instructions. 

Although our work resembles these works in the form of natural language instructions, we note several important differences. 
First, existing works focus on general instruction following that is applicable to a very broad range of tasks
and evaluate on generalization capabilities across tasks. We specifically consider the use of NL commands in CTG setting, and compare variants of CTG models in tailored test scenarios. Moreover, previous works in natural language instruction employ a fixed number of templates for each task, whereas we craft a PCFG that can generate a diverse set of command sentences to serve as templates. We show the effectiveness of our PCFG over fixed-set templates in subsequent experiments in Section \ref{exp:pcfg_generalize}. Finally, prompting models with NL instructions fails for moderately sized LMs without any modifications \citet{li-liang-2021-prefix}. Thus, it is non-trivial to adapt NL instruction to smaller models.



\paragraph{Controlled Text Generation} 

In open-ended text generation, a series of approaches have been proposed to control the generation to satisfy certain attributes (e.g. topic) \citep[\textit{inter alia}]{nitish2019ctrl,Dathathri2020Plug,krause-etal-2021-gedi-generative,yang-klein-2021-fudge,liu-etal-2021-dexperts}. Some of these studies utilize a trained classifier to guide the generative model towards the desired attribute, while others use a smaller LM to reweight LM logits. Very recently, \citet{Li2022DiffusionLMIC} focus on controlling more complex attributes such as syntactic structure with a non-autoregressive LM. Another line of work conduct CTG via prompt learning \citep{DBLP:journals/corr/abs-2110-08329, Yang2022TailorAP}. These work assume a fixed set of control attributes. 

Our NL interface is more related to \citet{yu-etal-2021-attribute-alignment}, which uses an attribute alignment function to embed attribute words into a hidden representation that guides LM generation. The attribute alignment function does not assume attribute tokens are from a fixed set, so it is possible to do inference on an attribute token not seen in training. Keyword2Text \citep{pascual-etal-2021-plug-play} shift the distribution over vocabulary toward words that are semantically similar to control keywords in a discriminator-free manner, thus does not assume a fix set of keywords. Besides attribute control, lexically constrained decoding \citep{post-vilar-2018-fast} has also been used to enforce certain key phrases to be included in the generation \citep{yuning2020constrainedsummary}. Other types of control includes conditioning LM output on semantic signatures \citep{weir-etal-2020-cod3s}. Recently, \citet{Qin2022COLDDE} unifies hard constraints (e.g., keyword appearance) and soft semantic constraints through an energy function.

\section{Framework}


The goal of controlled text generation is to model the conditional distribution $P(x|a)$ so that the generated text $x$ satisfies the desired attributes $a$. $a$ could include multiple attributes (e.g., topic and length), and we will use $a_i$ to denote the $i$th attribute. In the standard categorical setting, the attribute $a_i$ are from a fixed set of pre-defined options. We assume there are $m$ attributes of interest ($m \leq 2$ in our experiments).
In the next few sections, we describe the PCFG which we craft to embed the categorical attributes, and our proposed NL variants of several existing CTG systems.

\subsection{Embedding Attributes into Commands}

We embed categorical attributes into natural language commands with a PCFG.\footnote{Note that our command generation process is not strictly a PCFG, but it is very close.} We favor PCFG due to its ability to generate diverse NL variations expressing the same control semantic. For simplicity, most of the probability weights are set to uniform. In this section we will describe it in high-level, and more details and the full set of rules are provided in \autoref{sec:app_pcfg}. 
\autoref{tab:pcfg_example} is a concrete example of how a command describing an AG news article with a sports topic could be generated by our PCFG. We clarify that while the PCFG is used for training and testing in our work, the end user won't need to use it, as the model can generalize to unseen commands (Section \ref{exp:pcfg_generalize}).


Our command generation has three steps. First, a template with $m$ attribute slots is generated by the PCFG. We design the PCFG to generate templates that ``ask'' the system to generate text with some attributes and domains. We first sample a top level seed template from \texttt{ROOT} that determines high level sentence structure (e.g., \texttt{[PLS]} \texttt{[HEAD-FORM]} a \texttt{[TEXT-FORM]} \texttt{[LABEL-SEG]}), then fill in sentence segments with PCFG rules (e.g., \texttt{[HEAD-FORM]} will be substituted by ``generate''). These sentence segments are neither domain nor attribute specific thus can be used regardless of the attributes. 
In contrast to writing a set of fixed templates, our PCFG has multiple levels of rule and can greatly improve NL variation.

Next, we verbalize the domain media $D$, attribute $a$, and attribute name $A$ into natural language by crafting PCFG rules that transform them into words or phrases. Considering the fact that different words could have similar meaning in natural langauge, these mappings could be one-to-many to further improve NL variation. For instance, news about ``business'' can also be described as ``commerce'', and ``very negative'' is similar to ``terrible''. 

Finally, we conduct a postprocessing step to correct simple grammar errors, e.g., ``a AG news article'' would be corrected as ``an AG news article''.

\begin{table}[]
\footnotesize
\begin{tabular}{l}
\toprule
\textbf{1. PCFG-based template generation}\\
(1) {Generate top-level seed template from} \texttt{ROOT}:\\
$\Rightarrow$ \textcolor{red}{\texttt{[PLS]} \texttt{[HEAD-FORM]} a \texttt{[TEXT-FORM]} \texttt{[LABEL-SEG]}.} \\

(2) {Select PCFG rules to generate template}: \\
\texttt{[PLS]} $\rightarrow$ \dots $\rightarrow$ please, \texttt{[HEAD-FORM]} $\rightarrow$ \dots $\rightarrow$ generate, \\
\texttt{[TEXT-FORM]} $\rightarrow$ \dots $\rightarrow$ $D$\\
\texttt{[LABEL-SEG]} $\rightarrow$ \dots $\rightarrow$ with a $a$ $A$\\
$\Rightarrow$ \textcolor{red}{please} \textcolor{red}{generate} a \textcolor{red}{$D$} \textcolor{red}{with a $a$ $A$}. \\
\midrule
\textbf{2. Verbalize} \\
$\Rightarrow$ please generate a \textcolor{red}{AG news report} with a \textcolor{red}{sports topic}. \\
\midrule
\textbf{3. Postprocess} \\
$\Rightarrow$ \textcolor{red}{Please} generate \textcolor{red}{an} AG news report with a sports topic. \\
\bottomrule
\end{tabular}
\caption{Examples of PCFG command generation. \texttt{ROOT} is the PCFG start symbol. Newly replaced segments are highlighted in red. In step 1.(2), we omit intermediate PCFG expansions to ``$\rightarrow$ \dots $\rightarrow$''.}
\label{tab:pcfg_example}
\end{table}

In our preliminary attempts, we attempted to train a conditional neural LM for command generation, instead of using a PCFG. Although the neural model has better diversity, the stochastic nature of sampling makes the attribute embedding inaccurate. Besides, training such a neural LM would require a large amount of (attribute, command) paired data.
Therefore we turn to a PCFG approach as it has guaranteed accuracy, with decent diversity.

\vspace{-1mm}
\subsection{Models}
\label{sec:models}


In this section we first review some existing CTG models. For the new NL interface we propose natural variants of the models which take commands as input. All models are based on a pretrained autoregressive LM, denoted by $P_\text{b}$.
\vspace{-1mm}
\subsubsection{PrefixLM}
A direct method to model the conditional distribution $P(x|a)$ is to encode the attribute as a prefix and finetune the base model to generate $x$ conditioned on the prefix. In the standard categorical attribute setting, we randomly initialize an embedding vector for each attribute, and feed the corresponding embeddings as prefix. Multiple attributes are arranged by a pre-defined order. 

\vspace{-1mm}
\paragraph{PrefixLM-NL}
The NL variant of PrefixLM is straightforward. We just use the command as prefix. No extra parameters need to be added.

\subsubsection{Future Discriminator Controlled Generation (FUDGE)}
FUDGE \citep{yang-klein-2021-fudge} decomposes the conditional distribution using Bayes' rule according to \autoref{eq:fudge-orig}:
\begin{equation}
\label{eq:fudge-orig}
    P_\text{fudge}(x_i|x_{1:i-1},a) \ \propto \  P_\text{b}(x_i|x_{1:i-1}) P_\text{cls}(a|x_{1:i}).
\end{equation}

It involves training a future discriminator to predict whether the generated prefix $x_{1:i}$ will lead to a full generation that satisfy the attribute $a$. Following FUDGE's original formulation, we assume different attributes are conditionally independent and train a discriminator $P(a_k|x_{1:i})$ for each attribute $a_k$. We then use their product as the probability that all attributes are satisfied, i.e.,  $P(a_1,\dots,a_m|x_{1:i})=\prod_kP(a_k|x_{1:i})$.

As we consider attributes with multiple options (e.g., 4 topics or 5 sentiments), the FUDGE discriminator for a single attribute is a multiclass classification model that predicts the conditional distribution $P(a|x_{1:i})$ over all possible options of attribute $a$.

\vspace{-1mm}
\paragraph{FUDGE-NL}


In order to enable FUDGE to handle natural-language commands, we utilize a binary alignment discriminator juding whether the generated text aligns with the command. Given a command $c$, let $y_c\in\{0,1\}$ be a binary variable that denotes whether the prefix $x_{1:i}$ aligns with the command. Control is achieved by generating from the conditional distribution $P(x_i|x_{1:i-1}, y_c=1)$ that the alignment property is satisfied. We modify FUDGE's decomposition as \autoref{eq:fudge-nl}:
\begin{equation}
\label{eq:fudge-nl}
    \begin{split}
        P_\text{fudge-nl}&(x_i|x_{1:i-1},y_c=1) \\
        &\propto \ P_\text{b}(x_i|x_{1:i-1})P_\text{cls}(y_c=1|x_{1:i}).
    \end{split}
\end{equation}

$P_\text{cls}(y_c=1|x_{1:i})$ is modeled by a binary classifier trained on a dataset of command and generation prefix pairs $\{(c, x_{1:i})\}$. To create this data, for a given example text $x$ with attributes $a$, we first apply our PCFG to generate a true command $c^\text{pos}$. We then randomly flip one (or both) of the attribute in $a$, and generate a false command $c^\text{neg}$. By pairing $c^\text{pos}$ and $c^\text{neg}$ with $x$, we obtain the positive/negative training data for the discriminator.
In practice, we concatenate the command and generation prefix (separated by a special \texttt{[SEP]} token), and feed it as input to the alignment discriminator.

\vspace{-1mm}
\paragraph{FUDGE-Binary}
One major difference between FUDGE and its NL variant is that the discriminator is always binary for FUDGE-NL due to the alignment objective. 
This inspires us to propose a binary variant of FUDGE model, FUDGE-Binary, which operates with the categorical interface. Similar to FUDGE-NL, we use a binary variable $y_a$ to denote whether $x_{1:i}$ aligns with attribute $a$, and modify the decomposition as:
\begin{equation}
\label{eq:fudge-binary}
    \begin{split}
        P_\text{fudge-bin}&(x_i|x_{1:i-1},y_a=1) \\ \propto \ &P_\text{b}(x_i|x_{1:i-1})P_\text{cls}(y_a=1|x_{1:i}).
    \end{split}
\end{equation}

FUDGE-Binary's discriminator will always makes a binary prediction even if there are more than two options for a single attribute. Since attributes are still from a fixed set, we use a single classification model but attach a separate classifier head for each option.
During training, the classification head $W_{a^*}$ that matches the correct attribute $a^*$ receives a correct label $y=1$, and all other classification heads $\{W_a\}_{a\neq a^*}$ receive label $y=0$. At test time, we select the classification head $W_a$ base on the desired attribute $a$ to predict the alignment probability $P(y_a=1|x_{1:i})$.
Although this variant is a simple modification from the original FUDGE, empirically we find it to achieve a stronger performance in the categorical interface.







\section{Experiments}


In this section, we first compare natural language CTG models with their categorical counterparts under full-data setting, and then design experiments to test natural language CTG models' generalization capabilities, where the models need to generalize to (1) unseen commands (2) unseen attribute options (3) unseen combinations of attribute options. We begin by describing our datasets and metrics.
\subsection{Datasets}

We utilize two popular text classification datasets for our experiments: AG News and Yelp Review.\footnote{Obtained from Hugging Face Datasets.} For each dataset, we consider two control attributes: label and length. The label attribute is extracted from the classification label, i.e., topic labels for AG News and sentiment labels for Yelp Review. There are 4 topics \{world, sports, business, science/tech\} in AG News and 5 sentiment classes ranging from most positive to most negative in Yelp Review. The length attribute is created by dividing the dataset to $n_\text{len}$ length ranges so that number of training examples in each length range is balanced. We use $n_\text{len}=3$ for AG News and $n_\text{len}=5$ for Yelp Review. We refer readers to \autoref{sec:app_dataset} for details about dataset preprocessing.


\subsection{Evaluation metrics}

\paragraph{Control Accuracy} For the main metric evaluating the effectiveness of the control, we consider three types of control accuracy: \textsc{Label Accuracy} refers to the accuracy that the generated text satisfies the classification label, i.e., topic classification accuracy on AG News and sentiment classification accuracy on Yelp. This metric is computed by a RoBERTa classifier fine-tuned on the corresponding classification dataset. \textsc{Length Accuracy} refers to the accuracy that the generated text's tokenized length lies within the predefined length range. \textsc{Compositional Accuracy} refers to the accuracy that both the label and length attributes are satisfied.


\vspace{-1mm}
\paragraph{Text Quality} We consider three metrics to measure the quality of the generated text. 
\textsc{GPT-Neo Perplexity (G-PPL)}: we finetune the GPT-Neo-1.3B model\footnote{A publicly-available replication of GPT-3 obtained from \url{https://huggingface.co/EleutherAI/gpt-neo-1.3B}} on the corresponding datasets (without the labels), and report the perplexity of the generated text given by it. \textsc{BLEU} score: we randomly sample 100 examples from the AG News or Yelp test set as the reference, and compute the 4-gram BLEU score. 
\vspace{-1mm}
\paragraph{Diversity} We measure diversity of the generated text using  4-gram \textsc{Text Entropy} \citep{Zhang2018GeneratingIA}. That is, treat the generated token frequency as a discrete distribution, and compute its entropy.

\subsection{Model instantiation}
Here we describe the implementation of models mentioned in Section \ref{sec:models}. We use the Hugging Face transformers library \citep{wolf-etal-2020-transformers} and adapt from FUDGE's released code.\footnote{Our code and data will be released in \url{https://github.com/jackjyzhang/pcfg-nl-interface}.}

For all models, we produce generation by top-$k$ sampling with $k=20$ unless otherwise stated.

\vspace{-1mm}
\paragraph{PrefixLM variants}
We finetune a GPT-2 \citep{gpt2_radford2019language} small model without any modification (except for adding necessary special tokens) for both PrefixLM and PrefixLM-NL. At test time, we feed the desired attributes or command sentence as a prefix, and evaluate on the continuation the model produced.

\vspace{-1mm}
\paragraph{FUDGE variants} 

The backbone language model $P_b$ for FUDGE models is a GPT-2 small model finetuned on the corresponding dataset, using the same data available at discriminator training. That is, under zero-shot setting, we use the same data configuration to finetune the backbone LM.

For FUDGE and FUDGE-Binary, we train two discriminator for each of the label (topic or sentiment) and length attribute; FUDGE-NL use a single alignment discriminator to handle commands.

Each discriminator for FUDGE and FUDGE-NL is a GPT-2 small model followed by a single linear classification layer (with different number of output classes).
The discriminator for FUDGE-Binary is a GPT-2 small model followed by multiple linear classification layers, with each one corresponds to an option for the label or length attribute. Each classification layer makes a binary prediction about whether the generation prefix satisfies the particular option of the attribute.



\subsection{Full-data Setting}
\label{sec:fulldata}

\begin{table*}[h]
\centering
\footnotesize
\begin{tabular}{clcccccc}
\toprule
 & & \textit{\textbf{Diversity}} & \multicolumn{2}{c}{\textit{\textbf{Text Quality}}} & \multicolumn{3}{c}{\textit{\textbf{Control Accuracy}}} \\
\cmidrule(lr){3-3} \cmidrule(lr){4-5} \cmidrule(lr){6-8} \textsc{Dataset} & \textsc{Method} & \textsc{Ent.} $\uparrow$ & \textsc{G-PPL} $\downarrow$ & \textsc{BLEU} $\uparrow$ & \textsc{Label} $\uparrow$ & \textsc{Length} $\uparrow$ & \textsc{Comp.} $\uparrow$ \\ \midrule
\multirow{6}{*}{AG News}
& PrefixLM & 11.325 & 11.369 &.862 &.907 &.559 &.574 \\
& PrefixLM-NL & 11.371 & 12.126 &.866 & \textbf{.933} & \textbf{.677} & \textbf{.612} \\
\cmidrule(lr){2-8}
& FUDGE & 11.286 & 12.055 &.862 &.963 &.962 &.880 \\
& FUDGE-Binary & 11.276 & 12.617 &.864 & \textbf{.980} &.958 & \textbf{.918} \\
& FUDGE-NL & 11.368 & 12.197 &.865 &.965 & \textbf{.972} &.914 \\
\midrule
\multirow{6}{*}{Yelp Review} 
& PrefixLM & 11.800 & 10.406 &.942 & \textbf{.644} & \textbf{.949} & \textbf{.590} \\
& PrefixLM-NL & 11.828 & 10.361 &.943 &.637 &.919 &.547 \\
\cmidrule(lr){2-8}
& FUDGE & 11.217 & 10.628 &.940 &.620 &.794 &.564 \\
& FUDGE-Binary & 11.727 & 10.402 & .943 & \textbf{.871} & .942 & \textbf{.805} \\
& FUDGE-NL & 11.802 & 10.410 &.941 &.775 & \textbf{.972} &.640 \\
\bottomrule
\end{tabular}
\caption{\label{tab:reg_results} Results for full-data setting. NL model performance is on par with their categorical counterparts.}
\end{table*}


In full-data setting, we train the models on all data of the AG News or Yelp review dataset, with the purpose to test whether the NL interface would degrade the model's performance. This is the regular set up for existing works on CTG except that we aim to control two attributes simultaneously instead of one. The results for full-data setting is shown in \autoref{tab:reg_results}, with qualitative examples available in \autoref{tab:full_data_examples_ag} and \autoref{tab:full_data_examples_yelp} in the appendix. 

\vspace{-1mm}
\paragraph{Performance comparison between the NL and categorical interface} We notice that the generated text quality and diversity between different models are similar in the full-data setting. While PrefixLM-NL and its categorical variant PrefixLM has similar control accuracy on both datasets, FUDGE-NL consistently outperforms the original FUDGE setup. In either case, the performance of the NL variant is on par with its original model, suggesting our NL interface does not degrade CTG performance in the full-data setting.

Somewhat surprisingly, FUDGE-Binary outperforms FUDGE-NL and the original FUDGE model, especially on the Yelp dataset where the classification is more difficult. The reason could be that the task of the binary classification is less noisy than the multiclass classification, which leads to stronger control.


\vspace{-1mm}
\paragraph{Performance across model families} 
Across two datasets, FUDGE-based models outperforms PrefixLM models, with the exception that FUDGE does not beat (but is comparable to) PrefixLM on Yelp. This is largely consistent with previous results that discriminator-based CTG approaches can achieve higher controllability than conditional LMs \citep{yang-klein-2021-fudge, krause-etal-2021-gedi-generative}. 


\subsection{Generalizing to Unseen Commands}
\label{exp:pcfg_generalize}

\begin{table*}[h]
\centering
\footnotesize
\begin{tabular}{clcccccc}
\toprule
 & & \textit{\textbf{Diversity}} & \multicolumn{2}{c}{\textit{\textbf{Text Quality}}} & \multicolumn{3}{c}{\textit{\textbf{Control Accuracy}}} \\
\cmidrule(lr){3-3} \cmidrule(lr){4-5} \cmidrule(lr){6-8} \textsc{Dataset} & \textsc{Method} & \textsc{Ent.} $\uparrow$ & \textsc{G-PPL} $\downarrow$ & \textsc{BLEU} $\uparrow$ & \textsc{Label} $\uparrow$ & \textsc{Length} $\uparrow$ & \textsc{Comp.} $\uparrow$ \\ \midrule
\multirow{6}{*}{AG News}
& PrefixLM-NL-T20 & 11.412 & 12.345 & .865 & .922 & .522 & .458 \\
& PrefixLM-NL-T40 & 11.405 & 11.981 & .863 & .923 & .496 & .424 \\
& PrefixLM-NL-PCFG & 11.381 & 12.350 & .868 & \textbf{.933} & \textbf{.567} & \textbf{.505} \\
\cmidrule(lr){2-8}
& FUDGE-NL-T20 & 11.368 & 11.677 & .864 & .936 & .717 & .603 \\
& FUDGE-NL-T40 & 11.355 & 11.678 & .864 & .938 & .759 & .664 \\
& FUDGE-NL-PCFG & 11.369 & 12.174 & .863 & \textbf{.955} & \textbf{.936} & \textbf{.826} \\
\midrule
\multirow{6}{*}{Yelp Review} 
& PrefixLM-NL-T20 & 11.916 & 10.523 & .943 & .389 & .612 & .177 \\
& PrefixLM-NL-T40 & 11.935 & 10.309 & .943 & .398 & .603 & .216 \\
& PrefixLM-NL-PCFG & 11.869 & 10.251 & .945 & \textbf{.443} & \textbf{.721} & \textbf{.250} \\
\cmidrule(lr){2-8}
& FUDGE-NL-T20 & 12.155 & 9.567 & .936 & .364 & .531 & .148 \\
& FUDGE-NL-T40 & 11.918 & 9.986 & .944 & .538 & .619 & .249 \\
& FUDGE-NL-PCFG & 11.836 & 10.341 & .941 & \textbf{.687} & \textbf{.864} & \textbf{.462} \\
\bottomrule
\end{tabular}
\caption{\label{tab:pcfg_results} Results for experiment on PCFG effectiveness. Training natural language CTG models with PCFG-generated commands greatly improves controllabilty on unseen commands, compared to models trained on fixed-set templates.}
\end{table*}

In this section, we design a set of experiments to test natural language CTG model’s ability to generalize to commands unseen during training. Specifically, we compare the effectiveness of our proposed PCFG with commands generated by fix-set templates, as adopted in previous works \citep{Sanh2021MultitaskPT, wei2022finetuned, Mishra2022CrossTaskGV}.

To create a setup similar to previous work, we hand-crafted 20 diverse templates for each dataset. This is already twice the number of templates used in \citet{wei2022finetuned} and comparable to the number of seed templates in our PCFG. We denote models trained on this set of templates by ``-T20'' suffix. We also explore a stronger version of fix-set template by doubling the number of templates, totalling 40 templates for each dataset, denoted by ``-T40'' suffix.
We test the above models on 20 hand-crafted unseen templates that are different from both the PCFG and fixed-set templates, and compare results with our proposed PCFG-based models, denoted by ``-PCFG'' suffix.

Results are shown in \autoref{tab:pcfg_results}. The results show that when conditioning on unseen commands, both the PrefixLM-NL and FUDGE-NL models that used PCFG has notably better controllability compared to fixed-set template models. Thus, the above experiments provide empirical evidence that \textbf{our PCFG can effectively improve the model's generalization ability on natural language variation within commands}.


\subsection{Generalizing to Unseen Attributes}

\begin{table*}[t]
\centering
\footnotesize

\begin{tabular}{clcccccccc}
\toprule
\multicolumn{1}{c}{} &  & \multicolumn{2}{c}{\textit{\textbf{Diversity}}} & \multicolumn{4}{c}{\textit{\textbf{Text Quality}}} & \multicolumn{2}{c}{\textit{\textbf{Control}}} \\ \cmidrule(lr){3-4} \cmidrule(lr){5-8} \cmidrule(lr){9-10}
\multicolumn{1}{c}{} &  & \multicolumn{2}{c}{\textsc{Ent.} $\uparrow$} & \multicolumn{2}{c}{\textsc{G-PPL} $\downarrow$} & \multicolumn{2}{c}{\textsc{BLEU} $\uparrow$} & \multicolumn{2}{c}{\textsc{Acc.} $\uparrow$} \\
\midrule
\textsc{Setup} & \textsc{Method} & Z.S. & Reg. & Z.S. & Reg. & Z.S. & Reg. & Z.S. & Reg. \\ \midrule
\multirow{1}{*}{No Control Baseline} 
 & GPT-2-finetuned & 9.745 & 9.735 & 11.050 & 11.062 & .866 & .867 & .009 & .343 \\
\midrule
\multirow{4.5}{*}{Zero-shot data} 
 & PrefixLM-NL & 9.736 & 9.726 & 14.797 & 11.556 &.867 &.860 & \textbf{.222} & \textbf{.967} \\
 & FUDGE-NL & 9.359 & 9.748 & 21.604 & 11.497 &.601 &.863 &.038 &.927 \\
 \cmidrule(lr){2-10}
 & PrefixLM-NL-unb & 9.738 & 9.737 & 12.980 & 11.387 &.871 &.862 & \textbf{.204} & \textbf{.913} \\
 & FUDGE-NL-unb & 9.537 & 9.762 & 21.547 & 11.795 &.623 &.862 &.203 &.773 \\ \midrule
\multirow{4.5}{*}{+Extra data} 
 & PrefixLM-NL & 9.772 & 9.759 & 17.559 & 12.521 &.868 &.860 & \textbf{.448} & \textbf{.960} \\
 & FUDGE-NL & 9.536 & 9.741 & 22.727 & 11.430 &.782 &.863 &.071 &.935 \\
  \cmidrule(lr){2-10}
 & PrefixLM-NL-unb & 9.734 & 9.752 & 14.611 & 11.716 &.867 &.861 & \textbf{.455} & \textbf{.928} \\
 & FUDGE-NL-unb & 9.587 & 9.748 & 24.898 & 11.933 &.769 &.864 &.416 &.784 \\ \bottomrule
\end{tabular}

\caption{\label{tab:zs_results} Results for zero-shot setting. \textsc{Z.S.} (zero-shot) denote metrics computed with the zero-shot class, \textsc{Reg.} (regular) denote metrics computed with seen classes during training. The simple PrefixLM-NL approach outperforms FUDGE-NL. Adding extra data doubles the zero-shot accuracy.} 
\end{table*}

\begin{table*}[t]
\centering
\footnotesize
\begin{tabular}{@{}rlcccccc@{}}
\toprule
 &  & \textit{\textbf{Diversity}} & \multicolumn{2}{c}{\textit{\textbf{Text Quality}}} & \multicolumn{3}{c}{\textit{\textbf{Compositional Accuracy}}} \\ 
\cmidrule(lr){3-3} \cmidrule(lr){4-5} \cmidrule(lr){6-8} \textsc{Dataset} & \textsc{Method} & \textsc{Ent.} $\uparrow$ & \textsc{G-PPL} $\downarrow$ & \textsc{BLEU} $\uparrow$ & \textsc{Test} $\uparrow$ & \textsc{Orig.} $\uparrow$ & \textsc{Diff.} $\downarrow$ \\ \midrule
\multirow{2}{*}{AG News} 
& PrefixLM-NL & 10.293 & 11.793 &.861 & \textbf{.593} &.612 & \textbf{.019} \\
& FUDGE-NL & 10.140 & 57.295 &.677 &.548 &.914 &.366 \\
\midrule
\multirow{2}{*}{Yelp Review} 
& PrefixLM-NL & 10.892 & 13.831 &.944 &\textbf{.537} &.547 & \textbf{.010} \\
& FUDGE-NL & 9.725 & 19.335 &.779 &.046 &.640 &.551 \\
\bottomrule
\end{tabular}
\caption{\label{tab:comp_results} Results for compositional setting. \textsc{Test} denote accuracy for unseen attribute combinations, \textsc{Orig.} denote accuracy in full-data setting, and \textsc{Diff.} shows the difference. PrefixLM-NL suffers little performance loss when generalizing to unseen attribute combinations, but FUDGE-NL's performance substantially degrades.}
\end{table*}

CTG models with categorical attributes can only control a fixed set of attribute options. {It is impossible for these models to control unseen attribute options without re-training due to architecture constraints} (e.g., FUDGE trains a classifier with a fixed number of labels). On the other hand, our proposed NL interface naturally allows CTG models to generalize to unseen options by embedding embedding novel attributes into a NL command using a verbalizer phrase unseen during training, as long as the novel attributes could be described in natural language. 
In this section, we conduct experiments to test our PCFG-based natural language CTG models' capabilities to generalize control to unseen attribute options. 

\vspace{-1mm}
\paragraph{Experimental setup} In this section, we control a single attribute (topic or sentiment) for ease of presentation. Although it is possible to also experiment on the length attribute, they are similar in nature. For an attribute with $n$ classes (e.g., 4 different topics or 5 different sentiments), we create $n$ zero-shot data splits and delete examples from one of the $n$ classes (i.e. the zero-shot class) completely during training. We test on both the zero-shot class and other seen classes separately, and report the average result over all $n$ splits. We conduct zero-shot experiments on the AG News dataset.

\vspace{-1mm}
\paragraph{Adding extra data} 
Since natural-language CTG models does not assume the attribute is from a fix set of options, it is possible to train the model to control attributes by using extra data with different attribute options. This is another capability enabled by our NL interface, previously unavailable due to the fix-set assumption. We experiment training the models on the zero-shot AG News split along with similar datasets in the news domain, aiming to test whether the model can learn from extra data and generalize to a wider range of attribute options. We utilize three extra news topic classification datasets: News Popularity, News Category, and the Inshorts News dataset.\footnote{Obtained from Hugging Face Datasets and Kaggle.} Topics that overlap with AG News are removed. We refer readers to \autoref{sec:app_dataset} for more details. For these datasets we use the same PCFG as AG News. When mixing multiple datasets during training, we follow \citet{JMLR:v21:20-074} and use examples-proportional mixing to control the relative frequency of examples from each dataset. We set the artificial limit of each extra dataset to the size of original AG News dataset.

The zero-shot results are shown in \autoref{tab:zs_results}. Since the categorical interface does not allow unseen category, we introduce a no-control baseline by finetuning the base LM with the same zero-shot data, and produce generations from it directly without control. Both FUDGE-NL and PrefixLM-NL beat this baseline. 

We observe that the simple PrefixLM-NL approach outperforms FUDGE-NL by a large margin in both zero-shot data and zero-shot + extra data setting. Moreover, as measured by both perplexity and BLEU, PrefixLM has higher generation quality as well. While there is still a large gap between the zero-shot and non-zero-shot label accuracy, \textbf{the extra data approach managed to \textit{double} the zero-shot accuracy in both NL models, 
showing the generalization potential of the natural language interface}. Qualitatively (shown in \autoref{tab:zero_shot_prefixlm} to \autoref{tab:zero_shot_fudge_ed}), we found that in cases where the output has the wrong topic, there are still signs that the generation is guided by the command. For example, when we zero-shot on the \textit{world} topic, we obtain text about sports with multiple country names.

\vspace{-1.5mm}
\paragraph{Backbone unblock experiment} Due to the nature of zero-shot experiment, we also block examples of the zero-shot class from the finetuning data of the backbone language model $P_\text{b}$. As a comparison, we try finetuning $P_\text{b}$ with full data, while still blocking the zero-shot class from prefix or classifier training, which mimic the setting where only unlabeled data is available. 


Results are shown in \autoref{tab:zs_results} as the ``-unb'' models. We observe a large performance boost for the FUDGE-NL model. This shows that extra unsupervised data is also helpful for control generalization. 

\subsection{Generalizing to Unseen Attribute Combinations}
\vspace{-1mm}

In this section we design experiments to test whether the models can generalize to unseen \textit{combinations} of attributes during training to test their compositional generalization abilities. In the following we describe our setup for AGNews, and it is similar for Yelp. 

Following \citet{Lake2018GeneralizationWS}, for each split, we select one of the topic classes (e.g., sports) as the non-compositional class, and for all training samples with this class, we do not include length in attributes or commands (i.e., the model never see combinations of \texttt{sports} and any length attribute in training). Note that the combinations of length attributes and other topics classes are kept (e.g., the model still see combinations of \texttt{business} and \texttt{short length}). At test time, we set the topic to be the non-compositional class and randomly sample the length attribute to control. We run experiments across all $n$ possible compositionality splits and report the averaged result.

Results are shown in \autoref{tab:comp_results}, with qualitative examples available in \autoref{tab:comp_ag_prefixlm} to \autoref{tab:comp_yelp_fudge}. We focus on the accuracy gap between this compositionality setting and the full-data setting. PrefixLM-NL has little trouble generalizing to unseen attribute combinations as indicated by the small gap. However, FUDGE-NL performed poorly on generalizing to unseen attribute combinations. Not only did FUDGE-NL's compositional accuracy dropped by a large margin, it also produced low quality text. 




\section{Conclusion}
In this work we propose a natural language interface for CTG, where we craft a PCFG to embed categorical attributes into natural language commands. We propose variants of existing CTG models that take commands as input. We first conduct experiments in the regular full-data setting, and find that our proposed NL variants of CTG models has comparable performance to the original ones.

We then design tailored experiments to test natural language CTG model's generalization capabilities. We show that our PCFG-based command generation approach is effective for handling unseen commands compared to fix-set templates. Additionally, our proposed NL models can effectively generalize to unseen attributes, an ability newly enabled by the NL interface. Finally, we find the simple PrefixLM approach shows robust generalization ability with the NL interface and outperforms FUDGE-based models, demonstrating significant modelling challenges and potentials with this new interface. We hope our work could motivate further research into this challenging interface for CTG.



\section{Limitations}
In this section, we point out several limitations restricted by the scope of our work. While the PCFG we create has decent diversity and is guaranteed to be accurate in embedding attributes, they are still rule-based and could not cover all the variations in natural language.

The natural language interface brings modelling challenges. The CTG model is now required to first extract salient information from the command sentence, while in the original categorical interface they are provided directly.

In this work, we have focused our experiments on PrefixLM and FUDGE. While these approaches are representative, there are still other relevant models we did not test. For instance, guiding the generation of an LM with a smaller LM \citep{liu-etal-2021-dexperts}, or prompt-based CTG approaches such as \citet{Yang2022TailorAP}. It would also be interesting to test how other models perform under the NL interface.


Finally, while we experiment controlling more than a single attribute in a single CTG model, in principle a NL command could be more complex and fine-grained. For example, it is possible to describe detailed semantic or syntactic constraints in a command sentence, and we leave those to future work.




\section*{Ethics Statement}
We acknowledge controlled text generation is potentially capable of generating harmful outputs such as producing offensive languages or hate speech. However, it is also shown in previous work that controlled text generation techniques can achieve text detoxification if used properly \citep{Dathathri2020Plug, krause-etal-2021-gedi-generative}. When changing the control interface from a categorical setting to natural language commands, we are giving the user a larger freedom of input. Thus, extra care should be taken when deploying natural-language controlled text generation models to the general public to avoid malicious user inputs.



\bibliography{anthology,custom}
\bibliographystyle{acl_natbib}

\clearpage
\appendix







\section{Dataset Details}
\label{sec:app_dataset}

\subsection{Main datasets}

\paragraph{Yelp Review} This is a dataset of user-written reviews for Yelp. It is a text classification dataset where the 5-sentiment labels are inferred from 1 to 5 stars given to the review. For each star, there are 130,000 training examples and 10,000 testing examples. In total, there are 650,000 training examples and 50,000 testing examples. We limit text length to 200 after tokenization. After this preprocessing step, there are 450,773 training and 34,620 testing examples, for a total of 485,393 examples. We sample a validation set from the train set with about the same size as the test set, and create a final dataset with 415,901/34,872/34,620 train/val/test examples.

The label attribute for Yelp Review is constructed from the 5 sentiment labels, which we verbalize as \{very negative, negative, neutral, positive, very positive\}. For the length attribute, we create 5 length classes \{very short, short, medium-length, long, very long\} with cut-offs 43,72,104,144 so that number of training examples in each length class is balanced. The dataset is obtained from \url{https://huggingface.co/datasets/yelp_review_full}.

\paragraph{AG News} This is a news topic classification dataset with 4 topics \{world, sports, business, science/tech\}. The news text used is the title and description. For each topic, there are 30,000 training examples and 1,900 testing examples, for a total of 120,000 training and 7,600 testing examples. We limit text length to 256 after tokenization. After this pre-processing step, there are 119,955 training and 7,599 testing examples, for a total of 127,554 examples. We sample a validation set from the train set with about 10\% of the original train set size, and create a final dataset with 107,959/11,996/7,599 train/val/test examples.

We use the topic labels as the label attribute, while adding alternative names for the labels. For the length attribute, we limit text length to 256. Because the text length in AG News is concentrated in a narrow range, we create 3 length classes \{short, medium, long\} with cut-offs 43 and 56 to make the number of training examples in each class balanced. The dataset is obtained from \url{https://huggingface.co/datasets/ag_news}.

\subsection{Extra data} 
\paragraph{News Category} The News Category dataset contains about 200K news headlines and short descriptions between 2012 and 2018 obtained from HuffPost. The advantage of this dataset is that it has a wide variety of topics, thus making the corresponding template very diverse. The list of topics and corresponding article counts is shown in \autoref{lst:newscat}. We remove topics that has overlap with AG News: \texttt{THE WORLDPOST, WORLDPOST, WORLD NEWS, SPORTS, BUSINESS, SCIENCE, TECH}. The dataset is obtained from \url{https://huggingface.co/datasets/Fraser/news-category-dataset}.

\lstinputlisting[caption={News Category dataset topics with corresponding number of examples.}, label={lst:newscat}, float=*, multicols=4]{newscat_classes}

\paragraph{News Popularity} The News Popularity in Multiple Social Media Platforms dataset is a dataset of social media sharing data of news articles about \texttt{economy}, \texttt{microsoft}, \texttt{obama}, and \texttt{palestine}. We use the concatenation of the \texttt{headline} and \texttt{short\_description} fields as the news text. The size of this dataset is around 93K. The dataset is obtained from \url{https://huggingface.co/datasets/newspop}.

\paragraph{Inshort News}
The Inshort News dataset is a dataset of news with topics \texttt{sports}, \texttt{politics}, \texttt{entertainment}, \texttt{world}, \texttt{automobile}, and \texttt{science}. We remove the topics that has overlap with AG News: \texttt{sports}, \texttt{world}, \texttt{science}. The filtered dataset contains about 5K examples. The dataset is obtained from \url{https://www.kaggle.com/datasets/kishanyadav/inshort-news}.

\section{Experiment Details}
\subsection{Training}
On AG News, we use an Adam optimizer with a learning rate $0.00005$ and train 10 epochs to train the PrefixLM models as well as FUDGE discriminators. On Yelp Review, we use an Adam optimizer with a learning rate of $0.0001$ and train 5 epochs. We conduct all experiments on a single NVIDIA Tesla V100 GPU with 32GB memory. The training time of each model depends on the particular setup, but is within 24 hours for all models. The number of trainable parameters for the PrefixLM, PrefixLM-NL, and FUDGE-NL model is approximately 120M.

The number of trainable parameters for FUDGE and FUDGE-Binary is approximately 120M for each of label or length attribute model, and approximately 240M in total.

The FUDGE models have an extra backbone language model that is kept frozen during discriminator training. The size of this backbone language model is approximately 120M. Backbones are first fine-tuned on corresponding classification datasets with a learning rate of $0.0001$ for 5 epochs.

\subsection{Hyperparameter choice under different settings}
We find that the experimental results are not particularly sensitive to training hyperparameters such as learning rate and batch size. At testing, the FUDGE conditioning strength hyperparameter $\lambda$ does have a notable effect on control accuracy. We report results with $\lambda$ that gives the highest control accuracy while maintaining text quality. For the FUDGE model family (FUDGE, FUDGE-Binary, FUDGE-NL), we set $\lambda=14$ on the full-data and low-resource experiments, and $\lambda=6$ on zero-shot experiments. On compositionality experiments, we set $\lambda=6$ for AG News and $\lambda=4$ for Yelp Review. We set a smaller $\lambda$ for zero-shot and compositionality settings because a larger $\lambda$ in these cases leads to a significant increase in repetition. Following FUDGE's original setup, we consider only the top 200 possible output tokens when modifying the LM logits for computational efficiency.

\section{Command PCFG Details}
\label{sec:app_template}
\label{sec:app_pcfg}
The full template for the AG News and Yelp Review datasets are available in \autoref{lst:agnews} and \autoref{lst:yelp}. We briefly explain important elements of the custom PCFG syntax below:
\begin{itemize}
    \item We first randomly sample a template in the \texttt{<templates>} section. These are templates with attribute slots which will be filled later. Besides attribute slots, there are other nonterminals in the template that corresponds to sentence segments. Rules for these elements are written in the \texttt{<variables>} sections. 
    \item Rules in the \texttt{<variables>} sections are compressed PCFG where rules with the same LHS are grouped together in a single line. They constitute the verbalization of domain names, attribute names, as well as a variety of sentence segments to increase the diversity of the PCFG.
    \item To verbalize the label attribute, the \texttt{<label>} section contains the mapping from categorical class indices to verbalized class names. Since the mapping could be one-to-many, different verbalizations of the same attribute class is separated by a comma.
    \item To verbalize the length attribute, the \texttt{<length>} section contains length cut-off values with the corresponding verbalized length level names, having similar syntax with the \texttt{<label>} section. An example with tokenized length $l$ will be treated as the longest length level such that the corresponding cut-off does not exceed $l$.
\end{itemize}

\begin{lstlisting}[float=*, caption={PCFG template for AG News}, label={lst:agnews}]
<variables>
[TEXT-CLASS]	AG news, AG news
[TEXT-FORM]	[TEXT-CLASS], [TEXT-CLASS], [TEXT-CLASS] article, piece of [TEXT-CLASS], [TEXT-CLASS] report, [TEXT-CLASS] item, AG newspaper article
[HEAD-FORM]	give me, generate, tell me about, show, show me, fetch me, output, I need, I want, need, I request, write
[TOPIC-NOUN]	topic, topic, theme, focus
[TOPIC-NOUNED]	topic, topic, themed, focused, related
[TOPIC-PREP]	about, related to, concerning, regarding, pertinent to
[TOPIC-UPDATEWORD]	updated, informed
[TOPIC-SEG]	[TOPIC-PREP] [TOPIC], [TOPIC-PREP] [TOPIC], that is [TOPIC-PREP] [TOPIC], that is [TOPIC-PREP] [TOPIC], that can keep me [TOPIC-UPDATEWORD] with [TOPIC]
[TOPIC-BESEG]	[TOPIC-PREP] [TOPIC], [TOPIC-PREP] [TOPIC], [TOPIC-PREP] [TOPIC], can keep me [TOPIC-UPDATEWORD] with [TOPIC]
[PLS]	please, , 
[COMMA-PLS]	/ please, , # use '/' as comma (escaped)
[BEFORE-BE]	let it, make sure to, I want it to

<length>
43	short, concise, very short, pretty short, extremely short, extra short
56	medium-length, normal-length
256	long, lengthy, very long, pretty long, extremely long, extra long

<label>	[TOPIC] 
0	the world, the world, the globe, international matters
1	sports, sports, sporting events
2	business, business, commerce
3	science, science, technology, technology, tech

<templates>
# label and length
[HEAD-FORM] a [LENGTH] [TEXT-FORM] [TOPIC-SEG] [COMMA-PLS] .
[PLS] [HEAD-FORM] a [TEXT-FORM] . [BEFORE-BE] be [LENGTH] and [TOPIC-BESEG] .
[PLS] [HEAD-FORM] a [TEXT-FORM] [TOPIC-SEG], and I need it to be [LENGTH] .
[HEAD-FORM] a [TEXT-FORM] [TOPIC-SEG] , and [BEFORE-BE] be [LENGTH] [COMMA-PLS] .
[HEAD-FORM] a [TEXT-FORM] . I want the [TOPIC-NOUN] to be [TOPIC], and length to be [LENGTH] .
[HEAD-FORM] a [TEXT-FORM] . I want the length to be [LENGTH], and [TOPIC-NOUN] to be [TOPIC] .
[HEAD-FORM] a [TEXT-FORM] . [BEFORE-BE] be not only [LENGTH] but also have a [TOPIC-NOUN] on [TOPIC] .
# label only
[HEAD-FORM] a [TOPIC] [TOPIC-NOUNED] [TEXT-FORM] [COMMA-PLS] .
[PLS] [HEAD-FORM] a [TOPIC] [TOPIC-NOUNED] [TEXT-FORM] .
[HEAD-FORM] a [TEXT-FORM] [TOPIC-SEG] [COMMA-PLS] .
[PLS] [HEAD-FORM] a [TEXT-FORM] [TOPIC-SEG] .
[PLS] [HEAD-FORM] a [TEXT-FORM] . Let it have a [TOPIC] [TOPIC-NOUN] .
[HEAD-FORM] a [TEXT-FORM] . Let it have a [TOPIC] [TOPIC-NOUN] [COMMA-PLS] .
[HEAD-FORM] a [TEXT-FORM] . I want the [TOPIC-NOUN] to be [TOPIC] .
# length only
[HEAD-FORM] a [LENGTH] [TEXT-FORM] [COMMA-PLS] .
[PLS] [HEAD-FORM] a [TEXT-FORM] . [BEFORE-BE] be [LENGTH] .
[PLS] [HEAD-FORM] a [TEXT-FORM] , and I need it to be [LENGTH] .
[HEAD-FORM] a [TEXT-FORM] , and [BEFORE-BE] be [LENGTH] [COMMA-PLS] .
[PLS] [HEAD-FORM] a [TEXT-FORM] , and [BEFORE-BE] be [LENGTH] .
[HEAD-FORM] a [TEXT-FORM]. I want the length to be [LENGTH] .
[HEAD-FORM] a [TEXT-FORM]. I want the length to be [LENGTH] [COMMA-PLS] .
\end{lstlisting}

\begin{lstlisting}[float=*, caption={PCFG template for Yelp Review}, label={lst:yelp}]
<variables>
[TEXT-CLASS]	yelp review, yelp review, yelp comment
[TEXT-FORM]	[TEXT-CLASS], [TEXT-CLASS], [TEXT-CLASS] article, [TEXT-CLASS] passage, [TEXT-CLASS] paragraph, [TEXT-CLASS] piece, piece of [TEXT-CLASS], yelp review chapter, [TEXT-CLASS] item
[HEAD-FORM]	give me, generate, tell me about, show, show me, fetch me, output, I need, I want, need, I request, write
[SENT-NOUN]	tone, sentiment, attitude, mood
[SENT-PREP]	with, with, with, that has, / which has, of
[SENT-SEG]	[SENT-PREP] a [SENT] [SENT-NOUN]
[PLS]	please, , 
[COMMA-PLS]	/ please, , # use '/' as comma (escaped)
[BEFORE-BE]	let it, make sure to, I want it to

<length>
43	very short, pretty short, extremely short, extra short
72	short, concise
104	medium-length, normal-length
144	long, lengthy
200	very long, pretty long, extremely long, extra long

<label>	[SENT]
0	very negative, terrible, very bad, extremely negative
1	negative, bad
2	neutral, unopinionated
3	positive, good, promising
4	very positive, very good, excellent, splendid, extremely positive

<templates>
# label and length
[HEAD-FORM] a [LENGTH] [TEXT-FORM] [SENT-SEG] [COMMA-PLS] .
[PLS] [HEAD-FORM] a [TEXT-FORM] . [BEFORE-BE] be [LENGTH] and having a [SENT] [SENT-NOUN] .
[PLS] [HEAD-FORM] a [TEXT-FORM] [SENT-SEG], and I need it to be [LENGTH] .
[HEAD-FORM] a [TEXT-FORM] [SENT-SEG] , and [BEFORE-BE] be [LENGTH] [COMMA-PLS] .
[HEAD-FORM] a [TEXT-FORM] . I want the [SENT-NOUN] to be [SENT], and length to be [LENGTH] .
[HEAD-FORM] a [TEXT-FORM] . I want the length to be [LENGTH], and [SENT-NOUN] to be [SENT] .
[HEAD-FORM] a [TEXT-FORM] . [BEFORE-BE] be not only [LENGTH] but also have a [SENT] [SENT-NOUN] .
# label only
[HEAD-FORM] a [SENT] [SENT-NOUN] [TEXT-FORM] [COMMA-PLS] .
[PLS] [HEAD-FORM] a [SENT] [SENT-NOUN] [TEXT-FORM] .
[HEAD-FORM] a [TEXT-FORM] [SENT-SEG] [COMMA-PLS] .
[PLS] [HEAD-FORM] a [TEXT-FORM] [SENT-SEG] .
[PLS] [HEAD-FORM] a [TEXT-FORM] . Let it have a [SENT] [SENT-NOUN] .
[HEAD-FORM] a [TEXT-FORM] . Let it have a [SENT] [SENT-NOUN] [COMMA-PLS] .
[HEAD-FORM] a [TEXT-FORM] . I want the [SENT-NOUN] to be [SENT] .
# length only
[HEAD-FORM] a [LENGTH] [TEXT-FORM] [COMMA-PLS] .
[PLS] [HEAD-FORM] a [TEXT-FORM] . [BEFORE-BE] be [LENGTH] .
[PLS] [HEAD-FORM] a [TEXT-FORM] , and I need it to be [LENGTH] .
[HEAD-FORM] a [TEXT-FORM] , and [BEFORE-BE] be [LENGTH] [COMMA-PLS] .
[PLS] [HEAD-FORM] a [TEXT-FORM] , and [BEFORE-BE] be [LENGTH] .
[HEAD-FORM] a [TEXT-FORM]. I want the length to be [LENGTH] .
[HEAD-FORM] a [TEXT-FORM]. I want the length to be [LENGTH] [COMMA-PLS] .
\end{lstlisting}

\section{Qualitative Examples}
\label{sec:app_qual}
We show qualitative examples for different experimental settings in \autoref{tab:full_data_examples_ag} to \autoref{tab:comp_yelp_fudge}.

\begin{table}[]
\small
\begin{tabularx}{\linewidth}{X}
\toprule
\textbf{PrefixLM-NL: Show me an AG news. I want the length to be concise, and topic to be the world.} Two Indonesian women hostages released The two Indonesian women hostages have been released, two days before their scheduled arrival in the US under the terms of a new US policy in Southeast Asia. \\ \midrule
\textbf{FUDGE-NL: Generate an AG news article. Make sure to be not only pretty long but also have a focus on sporting events.} No. 18 Utah Defeats Oregon, 71-60 (AP) AP - Cedric Benson scored 29 points, Jeri Dean added 15 and 18 points, and No. 18 Utah defeated Oregon 71-60 on Saturday night in an exhibition game that featured a balanced offense and a balanced defense. \\ \midrule
\textbf{PrefixLM: [WORLD][SHORT]} Two Indonesian women hostages released Indonesia \#39;s Foreign Ministry said Wednesday two female Indonesian women hostages were released Tuesday at dawn from an Al-Arabiya TV station, a day after it broadcast a videotape of their captivity. \\ \midrule
\textbf{FUDGE: [TECH][LONG]} Macromedia \#39;s  \#39;Blog \#39; to Make New Web Sites Macromedia is setting out to make Web site-hosting Web sites, starting with blog, a key part of its strategy for making new sites more relevant and useful. \\ \midrule
\textbf{FUDGE-Binary: [BUSINESS][MEDIUM-LEN]} Update 2: Oil Prices Drop Below \$52 a Barrel Despite Hurricane Oil futures prices dropped below \$52 on Friday, even though Hurricane Ivan, which was projected to hit the Gulf Coast late last week, was nonetheless expected to remain a drag on global oil supplies \\ 
\bottomrule
\end{tabularx}
\caption{AG News generation in full-data setting. All models can provide sensible generations with descent quality.}
\label{tab:full_data_examples_ag}
\end{table}

\begin{table}[]
\small
\begin{tabularx}{\linewidth}{X}
\toprule
\textbf{PrefixLM-NL: Show me a yelp review passage. I want the length to be pretty short, and tone to be very negative.} This place is a total joke. I called to place an order for delivery tonight and received a message with no answer. \\ \midrule
\textbf{FUDGE-NL: Generate a yelp review passage. Let it be long and having an extremely positive attitude.} OMG I don't know where i got my hair cut. We started with the salon manager. I have to say I am a little disappointed on this. The place is nice, staff nice and the haircut was good. My stylist was very knowledgable. She did an excellent job with the color and cut. I was so impressed. I am going back here for my next cut. The staff is nice and very helpful. I will give it one more try before I leave. The price was very reasonable for it being in such a nice shopping center. I was surprised at the quality. \\ \midrule
\textbf{PrefixLM: [NEUTRAL][MEDIUM-LEN]} Great for a quick meal but not really great. They do have a very good salad bar and the pizza is pretty good as well. If they had more seating and less of those annoying flies, I might have given them 4 stars. I am looking forward to trying more of their pizzas but they need more seating.  I think I'll still go back, but not on a regular basis. \\ \midrule
\textbf{FUDGE: [NEGATIVE][SHORT]} Disorganized and unorganized.  No water in the shower and no soap dispenser in the bathroom.  Very noisy when you come in during lunch hour.  Very disappointed and will not go back. \\ \midrule
\textbf{FUDGE-Binary: [POSITIVE][SHORT]} Great location!  The pizza's are huge and the crust was thin.  We had the sausage, pepperoni and sausage pizza and sausage was delicious.  If you're looking for a good pie to eat late night, this is definitely the spot. \\ 
\bottomrule
\end{tabularx}
\caption{Yelp Review generation in full-data setting.}
\label{tab:full_data_examples_yelp}
\end{table}

\begin{table}[]
\small
\begin{tabularx}{\linewidth}{X}
\toprule
\textbf{Show me an AG news that is about business.} China Stocks Set to Open Lower (AP) AP - China's consumer and investment markets are set to open relatively slackened this week due in part to lower oil prices, but stocks were slightly higher Monday  \#151; despite falling in recent weeks on the strength of the dollar and other key economic indicators. \\ \midrule
\textbf{Give me an AG news item that can keep me updated with international matters, please.} China \#39;s economy booming, but challenges persist China \#39;s economy is booming and it is struggling to meet growing economic challenges, but it faces tough foreign competition  quot;from within its own ranks quot;, China \#39;s top official said Sunday. \\ \midrule
\textbf{Show a sports related AG news report.} Update 1: Tennis: Masters Cup Semi Draws Three Aussies Aussies were forced to face three Australian players they will face in next month \#39;s semi-finals, a spokesman said Tuesday. \\ \midrule
\textbf{Show an AG newspaper article. I want the topic to be technology.} Google \#39;s initial public offering The initial public offering of the Internet search engine is expected to be much higher than Google Inc \#39;s initial public offering. \\ \bottomrule
\end{tabularx}
\caption{Zero-shot generation of the PrefixLM-NL model. Zero-shot class generation is shown. Interestingly, the model seems to be generating overlapping topics between zero-shot and non-zero-shot classes, e.g., the generated business news talks about China's stock market, which the model probably learned from the world topic that is not blocked.}
\label{tab:zero_shot_prefixlm}
\end{table}

\begin{table}[]
\small
\begin{tabularx}{\linewidth}{X}
\toprule
\textbf{Give me a technology topic AG news.} Microsoft to launch Xbox One console in India Microsoft is set to announce an Xbox One console in India on December 1. According to ET Now, the company will announce the launch on December 10. Microsoft announced the console on December 8. Its rival PlayStation 4 is also coming soon. \\ \midrule
\textbf{Show me an AG news item. Let it have a world topic, please.} India plans to launch second space station A second crew to the International Space Station will start service next year as India completes a five-year mission to the moon and comes within one day of meeting the International Space Station crew. \\ \midrule
\textbf{Give me a sporting events themed AG news, please.} Tennis: Oleg Tsunoda, Serena Williams Open in London (Reuters) Reuters - Two days before the third consecutive U.S. Open, tennis's biggest event, Nov. 8, will be staged on a hilltop overlooking London's \\ \midrule
\textbf{I request a commerce related AG news, please.} U.S. Pounds Oil, Prices Surge in Asia (AP) AP - The United States pounds oil as the world's largest petroleum consumer enters a pivotal year in growth, pushing prices for gasoline and heating oil to new lows in Asia and boosting the U.S. energy industry. \\ 
 \bottomrule
\end{tabularx}
\caption{Zero-shot generation of the PrefixLM-NL model trained with extra data. Zero-shot class generation is shown. In the tech generation, the Microsoft topic is from one of the extra datasets.}
\label{tab:zero_shot_prefixlm_ed}
\end{table}

\begin{table}[]
\small
\begin{tabularx}{\linewidth}{X}
\toprule
\textbf{Write an AG news item. I want the topic to be the globe.} Sports industry executives to be legal firms CEO's offices on businesses, companies, companies not business clients, companies executives, business travelers. \\ \midrule
\textbf{Give me a sporting events related AG news.} Fierce Fighting Roars Past a Home on the Wild Atlantic OAST LONDON - A series of strong Atlantic storms ripped through a British home on Tuesday, unleashing heavy rains, fierce winds and ferocious waves that washed away scores \\ \midrule
\textbf{Please generate a piece of AG news that is regarding commerce.} Oracle Wins Contingency Bid For PeopleSoft (AP) AP - Oracle Corp.'s Oracle Corp. has won its contract for PeopleSoft Inc.'s business software division, Oracle Corp. said Friday.\\ \midrule
\textbf{Show an AG newspaper article. Let it have a technology theme, please.} UN envoy urges Sudan resolution Sudanese authorities in Africa's top government urged Sudan to adopt a UN resolution on humanitarian crises in Africa in an effort to end fighting in the war-torn country. \\ \bottomrule
\end{tabularx}
\caption{Zero-shot generation of the FUDGE-NL model. Zero-shot class generation is shown. In many cases the model ignores the command, and goes on to generate text related to seen topics.}
\label{tab:zero_shot_fudge}
\end{table}

\begin{table}[]
\small
\begin{tabularx}{\linewidth}{X}
\toprule
\textbf{Please generate a piece of AG news that is regarding the world.} Paisley, Sharman hold India to 4 wickets for second innings (AFP) AFP - Shane Warne's unbeaten 141 against South Africa at the WACA was bowled out in the second and final innings as Australia held India to four wickets for a total of 141. \\ \midrule
\textbf{Generate an AG news. Let it have a sports theme.} Beltre Leads Red Sox Past Yankees 7-4 (AP) AP - Mark Buehrle pitched seven shutout innings of two-hit ball, and Boston clinched its eighth straight AL East crown, 7-4 over the New York Yankees on Saturday. \\ \midrule
\textbf{Give me a commerce related AG news.} Yahoo! Earnings Fall in First Quarter Yahoo! Inc.'s (YHOO.O: Quote, Profile, Research) quarterly earnings fell sharply, as the world \#39;s largest Internet advertising and search company reported narrower \\ \midrule
\textbf{Please show me an AG news. Let it have a technology focus.} Tech giants IBM profits rise tech giants IBM and Hewlett-Packard reported big gains yesterday as their technology businesses grew more profitable in August. Shares of Hewlett-Packard jumped as much as 2. \\ \bottomrule
\end{tabularx}
\caption{Zero-shot generation of the FUDGE-NL model trained with extra data. Zero-shot class generation is shown. Signs of the model being influenced by the command is evident from time to time, e.g., the world topic generation, although is actually sports-topic, contains many country names.}
\label{tab:zero_shot_fudge_ed}
\end{table}

\begin{table}[]
\small
\begin{tabularx}{\linewidth}{X}
\toprule
\textbf{Generate an AG news. Make sure to be long and about international matters.} US Forces Storm School in Mosul, 40 Dead  MOSUL, Iraq (Reuters) - U.S. tanks, aircraft and  artillery rumbled through a Mosul rebel-held school on  Saturday, 40 days after a suicide car bomb blast killed at  least 39 people outside a police academy in a town near Baghdad. \\ \midrule
\textbf{Need an AG news. I want it to be extra short and can keep me updated with sports.} Soccer: England win draw at Moldova England won their opening game against Moldova after defeating Poland 1-0 in a friendly draw at Frno, the Football Association said Thursday. \\ \midrule
\textbf{I want an AG news report. I want it to be extremely short and can keep me informed with business.} Oil Prices Surge to Record Highs on Ivan Crushing \$52 Oil prices jumped to multi-week highs yesterday on fears that Hurricane Ivan would hit the Southeast US Gulf Coast early next week. \\ \midrule
\textbf{Give me an extra long AG news article related to science.} Microsoft to Release Longhorn as Open Beta Microsoft announced it would release Longhorn in 2006, with features such as a new file system, new system management, and support for 64-bit extensions in addition to standard 64-bit capabilities. \\ \bottomrule
\end{tabularx}
\caption{Compositional generation of the PrefixLM-NL model on AG News. Non-compositional class generation is shown. The model generate fluent, on topic text without much performance degradation from full-data setup.}
\label{tab:comp_ag_prefixlm}
\end{table}

\begin{table}[]
\small
\begin{tabularx}{\linewidth}{X}
\toprule
\textbf{Generate an AG news article. I want the focus to be international matters, and length to be pretty short.} Italy arrests Italian businessman Italy yesterday arrested a businessman and a German national working for a major French and British industrial group and two others for connections to a major European political party scandal in Italy two years ago. \\ \midrule
\textbf{Give me a piece of AG news. I want the focus to be sporting events, and length to be long.} NFL Wrap: Brees, Brees Lead the Giants to the Playoffs Jamal Lewis ran for 147 yards and two touchdowns Saturday night to lead the Giants to playoff berth games Sunday at Philadelphia and Sunday at Atlanta and clinch Sunday \#39;s NFC title \&lt;b\&gt;...\&lt;/b\&gt; \\ \midrule
\textbf{Need an AG news article. Make sure to be not only medium-length but also have a focus on commerce.} Oil prices rebound sharply from record highs \&lt;b\&gt...\&lt;/b\&gt VIYUYYashko men took no charge as they trawl past what lookto     evidence indicate     Since arriving earlier earlier early today many children wandered past makeshift shelters and makeshift stalls. At least two dozen workers - many of them women - entered the battered warehouse, but as... \\ \midrule
\textbf{Generate an AG news article. I want the focus to be technology, and length to be pretty short.} Computer security chief steps down Computer security chief William Donaldson steps down today, the government said today. Mr Donaldson was leaving his department in his final days of the career. \\ \bottomrule
\end{tabularx}
\caption{Compositional generation of the FUDGE-NL model on AG News. Non-compositional class generation is shown. The text quality is notably lower than both PrefixLM's compositional generations (\autoref{tab:comp_ag_prefixlm}) as well as FUDGE-NL's generation in full-data setting (\autoref{tab:full_data_examples_ag}).}
\label{tab:comp_ag_fudge}
\end{table}

\begin{table}[]
\small
\begin{tabularx}{\linewidth}{X}
\toprule
\textbf{Write an extremely short yelp review passage of a very negative tone.} This place is a joke!!!!!! I called to place an order for delivery tonight and they quoted me \$50! And the food wasn't even good at all!!! \\ \midrule
\textbf{Write an extremely short yelp review passage of a negative tone.} This place is a total joke. I wish I had just gone to another place before coming here. The food was ok (but not really worth wasting \$40 on a meal) The service suck!!! \\ \midrule
\textbf{Generate a yelp review. Let it be concise and having an unopinionated attitude.} Not bad, not great (except for the wait for service and for the food being cold). However, the place is really small and is always crowded. The food is good (although a bit pricey) and the ambiance is relaxing. But again not great.\textbackslash nThe servers are nice and accommodating. \\ \midrule
\textbf{I need a yelp review item. I want the length to be extra short, and tone to be promising} Very happy with this hotel! Good deals, clean room and great service. The staff was very helpful. Will stay here again. \\ \midrule
\textbf{I need a yelp review passage. I want the mood to be very good, and length to be very long} I am a HUGE fan of this place! This is the type of place that I go to when I'm in the area! The food is fantastic! I have always had great service here. I usually go at dinner time on weekends but have yet to be disappointed. I love their wings! They are not your average wings. They come covered with a wonderful sauce and just a hint of garlic! They are always hot. I am a big guy though, so I tend to eat wings on the go. \textbackslash n\textbackslash nTheir pizza is amazing as well. I love that you can order a large size with no toppings, which is very convenient for me. I also love their ranch dressing, which is amazing as well. \textbackslash n\textbackslash nI always look forward to coming back to this place for some wings! \\ \bottomrule
\end{tabularx}
\caption{Compositional generation of the PrefixLM-NL model on Yelp Review. Non-compositional class generation is shown. Similar to scenario in AG News (\autoref{tab:comp_ag_prefixlm}), the model can generalize to unseen attributes combinations without much performance degradations.}
\label{tab:comp_yelp_prefixlm}
\end{table}

\begin{table}[]
\small
\begin{tabularx}{\linewidth}{X}
\toprule
\textbf{Show a yelp review. Let it be not only lengthy but also have an extremely negative tone.} Love this place.  Great coffee and delicious sandwiches, wraps, and salads.  The service is wonderful and everyone is helpful and friendly. \\ \midrule
\textbf{Write a yelp comment that has a bad sentiment, and make sure to be short.} I really wanted to love this place.  I love the concept.  The food.  Unfortunately, I was not impressed. \\ \midrule
\textbf{I need a yelp review paragraph. Make sure to be not only short but also have a neutral attitude.} It was ok. Great tasting bread with your sandwiches and cookies on your cookie crust to make it look like a fancy bakery with your cookies.... No wow... Really!?!?!?..... Never ever ever ever ever when ever I have been. Kind it kinda of defeats thinking about this at all. Service goes off without any notice. \\ \midrule
\textbf{I request a yelp review piece. I want the length to be extra long, and tone to be positive.} So much for the good reviews. My sister-in-law was going to give this place a try, but it seems a little overrated.\textbackslash n\textbackslash nI came for lunch with her, and ordered a half salad, soup, and sandwich combo. My salad came with the soup, but only half of it was actually soup! The soup was alright, but it came with a small piece of lettuce on top. The sandwich was decent.\textbackslash n\textbackslash nThe sandwich was nothing spectacular. I've had better for half the price! \\ \midrule
\textbf{I want a piece of yelp comment that has a very good mood, and make sure to be pretty long.} Always a great place. Food \& service always great \& prices are reasonable especially for the quality \& quantity they give you. Food comes out hot. My kids eat there every time \& are always happy with their meals. Prices have always been very reasonable for the quality \& quantity they give you. Owner is the man, \& he is the reason I come back to this place, \& I hope he is getting his act together soon! Give it a try \& please try them out for yourself!! You will leave happy \& full!! :) Thanks Chef! Keep hustling for quality \& quality food! Happy eating! Thanks Chef :) :)) :) Enjoy! :-) :) See ya! :) ;) :) :) :) :) :) :) ;) ;) \\ \bottomrule
\end{tabularx}
\caption{Compositional generation of the FUDGE-NL model on Yelp Review. Non-compositional class generation is shown. Text quality is notably low, with the model generates repetitive phrases or emoji from time to time.}
\label{tab:comp_yelp_fudge}
\end{table}


\end{document}